\documentclass[11pt]{article}
\usepackage[]{acl}

\usepackage{times}
\usepackage{latexsym}
\usepackage[T1]{fontenc}
\usepackage[utf8]{inputenc}
\usepackage{microtype}
\usepackage{booktabs}
\usepackage{multirow}
\usepackage{array}
\usepackage{graphicx}
\usepackage{amsmath}
\usepackage{amssymb}
\usepackage{url}
\usepackage{tikz}
\usetikzlibrary{arrows.meta,positioning,fit,backgrounds,calc}
\usepackage{listings}
\usepackage{etoolbox}
\usepackage{seqsplit}
\newcommand{\hashv}[1]{\texttt{\footnotesize\seqsplit{#1}}}

\providecommand{\nolinenumbers}{}
\AtBeginEnvironment{figure}{\nolinenumbers}
\AtBeginEnvironment{figure*}{\nolinenumbers}
\AtBeginEnvironment{table}{\nolinenumbers}
\AtBeginEnvironment{table*}{\nolinenumbers}

\newcommand{\m}{$-$}

\title{Do Small Models Use the Law You Give Them?\\
Measuring Context Use on a Bilingual Bangladesh Legal Benchmark}

\author{%
  \textbf{Moniruzzaman Mahadi\textsuperscript{1}, \ Abrar Mohammed Tanzim Alam\textsuperscript{1}, \ Sayma Siddika Monalisa\textsuperscript{1}} \\
  \textbf{Mir Mohammad Asif Abdullah\textsuperscript{1}, \ Swakkhar Shatabda\textsuperscript{2}, \ Md Adnan Arefeen\textsuperscript{1}} \\[5pt]
  {\normalsize \textsuperscript{1}North South University \qquad \textsuperscript{2}BRAC University} \\[5pt]
  {\normalsize \href{https://huggingface.co/datasets/momahadi/bangladesh-legal-qa-dataset}{\textcolor{blue}{HuggingFace/Datasets/Bangladesh-Legal-QA}}} \\[4pt]
  {\normalsize \textbf{Correspondence:} \texttt{momahadi9664@gmail.com}}}
  
\begin{document}
\maketitle

\begin{abstract}
Fine-tuning can improve legal question-answering accuracy without
improving how models use law supplied in context. We study
this distinction in bilingual Bangladeshi legal QA, where observed errors
can arise from answer scoring, retrieval, or failure to use relevant law.
We construct a hierarchy-preserving statutory corpus, 2,165 reviewed
bilingual fine-tuning examples, and a 150-item supplied-law control. We
evaluate six instruction-tuned models: Llama-3.2-1B, Llama-3.2-3B,
Qwen3.5-0.8B, Qwen3.5-2B, Qwen3.5-4B, and Gemma-4-E2B, with three
LoRA seeds per model. To separate effects, we combine constrained
option-letter scoring, cyclic option rotation, and controlled removal
of the governing provision. On 398 Bar Council outputs, an
exact-line parser attributes an accuracy gain of 50.0\% to the
Qwen3.5-2B seed-42 adapter, whereas option scoring yields only
3.0\%. For Gemma-4-E2B, the two scoring methods favor different systems.
When the governing provision is guaranteed to be present, five of six
reference models improve by 14.7\%--19.3\% under the four-order
criterion. Removing that provision reduces accuracy by
8.0\%--15.3\% for models and by 13.8\%--14.9\% points for their
adapters. However, difference-in-differences estimates show no increase in reliance on the governing provision after fine-tuning. Results show that legal adaptation claims require separating scorer,
retriever, and model effects. Our Code and data are available at
\url{https://anonymous.4open.science/r/bangladesh-legal-qa-11E3}.
\end{abstract}


\section{Introduction}
\begin{figure*}[t]
\centering
\definecolor{bxacc}{HTML}{1B3A5C}
\providecommand{\bxdet}{\scriptsize\color{black!62}}
\providecommand{\bxtick}[1]{\tikz[baseline=-0.55ex]{\draw[line width=0.7pt,
  draw=#1] (0,0) -- (0.075,-0.085) -- (0.245,0.155);}}
\begin{tikzpicture}[
  font=\small,
  every node/.append style={execute at begin node={%
    \hyphenpenalty=10000 \exhyphenpenalty=10000\relax}},
  panel/.style={fill=black!3, rounded corners=3pt, inner sep=2.5mm},
  db/.style={draw=black!35, fill=white, rounded corners=1.5pt, align=center,
             inner sep=3.5pt, text width=32mm, line width=0.4pt},
  dad/.style={db, draw=bxacc, line width=0.9pt},
  dev/.style={db, text width=19mm},
  head/.style={font=\footnotesize\bfseries, text=black!70},
  note/.style={font=\scriptsize, text=black!55, align=center},
  ar/.style={-{Latex[length=1.7mm,width=1.5mm]}, draw=black!50,
             line width=0.45pt}
]

\node[db, anchor=north] (src) at (1.85,3.55)
  {\textbf{six acts and three schedules}\\
   \bxdet parsed into hierarchy-preserving JSON};
\node[db, below=3mm of src] (gen)  {\textbf{3{,}425} candidate records};
\node[db, below=3mm of gen] (rev)
  {\textbf{author review, two passes}\\
   \bxdet every answer checked\\ against the source act};
\node[dad, below=6.5mm of rev] (keep)
  {\textbf{2{,}165 retained}\\
   \bxdet 1{,}211 Bangla and 954 English,\\
   each with its governing provision};

\draw[ar] (src)  -- (gen);
\draw[ar] (gen)  -- (rev);
\draw[ar] (rev) -- node[note, right=0.8mm] {1{,}260 cut} (keep);

\node[db, anchor=north] (ckpt) at (5.90,3.55)
  {\textbf{six models}\\
   \bxdet instruction-tuned, three families:\\
   Llama-3.2 1B and 3B;\\ Qwen3.5 0.8B, 2B and 4B;\\ Gemma-4-E2B};
\node[dad, below=12mm of ckpt] (lora)
  {\textbf{LoRA fine-tuning}\\
   \bxdet split 1{,}728 / 220 / 217,\\ disjoint by statutory group;\\
   seeds 17, 42 and 73};
\node[dad, below=6mm of lora] (sys)
  {\textbf{24 systems}\\
   \bxdet 6 reference and 18 adapters};

\draw[ar] (keep.east) -- ++(0.35,0) |- (lora.west);
\draw[ar] (ckpt.south) -- (lora.north);
\draw[ar] (lora.south) -- (sys.north);

\node[anchor=north] (tab) at (11.95,3.6) {%
  \setlength{\tabcolsep}{2.5pt}%
  \begin{tabular}{@{}l@{\hspace{1.5mm}}cccc@{\hspace{4mm}}cc@{}}
    & \multicolumn{4}{c}{\scriptsize\textbf{A. examination benchmark}}
    & \multicolumn{2}{c}{\scriptsize\textbf{B. control}}\\
    & \multicolumn{4}{c}{\scriptsize 398 rows (199 items $\times$ 2)}
    & \multicolumn{2}{c}{\scriptsize 150 items}\\
    \cmidrule(r){2-5}\cmidrule(l){6-7}
    & \scriptsize none & \scriptsize BM25 & \scriptsize dense
    & \scriptsize irrelevant & \scriptsize none & \scriptsize law\\
    \midrule
    \scriptsize reference
    & \bxtick{black!55} & \bxtick{black!55} & \bxtick{black!55}
    & \bxtick{black!55} & \bxtick{black!55} & \bxtick{black!55}\\
    \scriptsize\color{bxacc}\textbf{adapters}
    & \bxtick{bxacc} & \bxtick{bxacc} & \bxtick{bxacc}
    & \bxtick{bxacc} & \bxtick{bxacc} & \bxtick{bxacc}\\
    \bottomrule
  \end{tabular}};

\node[note, text width=70mm, below=6mm of tab] (ctx)
  {in \textbf{A} one prompt block holds no law, the top five BM25 hits, the top
   five dense hits, or length-matched irrelevant law;\\
   in \textbf{B} the governing provision sits among four same-act distractors,
   or is withheld};

\node[dev] (rot) at (11.95,0 |- sys) {\bxdet all four cyclic option rotations};
\node[dev] (argmax) at (9.4,0 |- rot)
  {\bxdet argmax over the\\ \texttt{A B C D} logits};
\node[dev] (agree)  at (14.5,0 |- rot) {\bxdet correct only if all four agree};

\draw[ar] (sys.east) -- (argmax.west);
\draw[ar] (argmax.east) -- (rot.west);
\draw[ar] (rot.east)    -- (agree.west);

\node[note, text width=70mm, below=2mm of rot] (foot)
  {reference and adapters read identical prompts, contexts and precision;
   adapter cells are the mean over three seeds};

\coordinate (t1) at (1.85,4.2);
\coordinate (t2) at (5.90,4.2);
\coordinate (t3) at (11.95,4.2);  \coordinate (b3) at (11.95,0 |- foot.south);
\begin{scope}[on background layer]
  \node[panel, fit=(src)(keep)(t1)] (p1) {};
  \node[panel, fit=(ckpt)(sys)(t2)] (p2) {};
  \node[panel, fit=(tab)(agree)(foot)(t3)(b3)] (p3) {};
\end{scope}

\node[head, below=1.5mm of p1.north] {THE RESOURCE};
\node[head, below=1.5mm of p2.north] {TRAINING};
\node[head, below=1.5mm of p3.north] {EVALUATION};
\end{tikzpicture}
\caption{Study design. Panel \textbf{B} places the governing provision in the
prompt, removing retrieval failure. The scoring chain then reads the
answer from the option-letter logits, removing format from the
measurement, and keeps it only if it survives all four option orders. A
tick marks each executed arm.}
\label{fig:design}
\end{figure*}
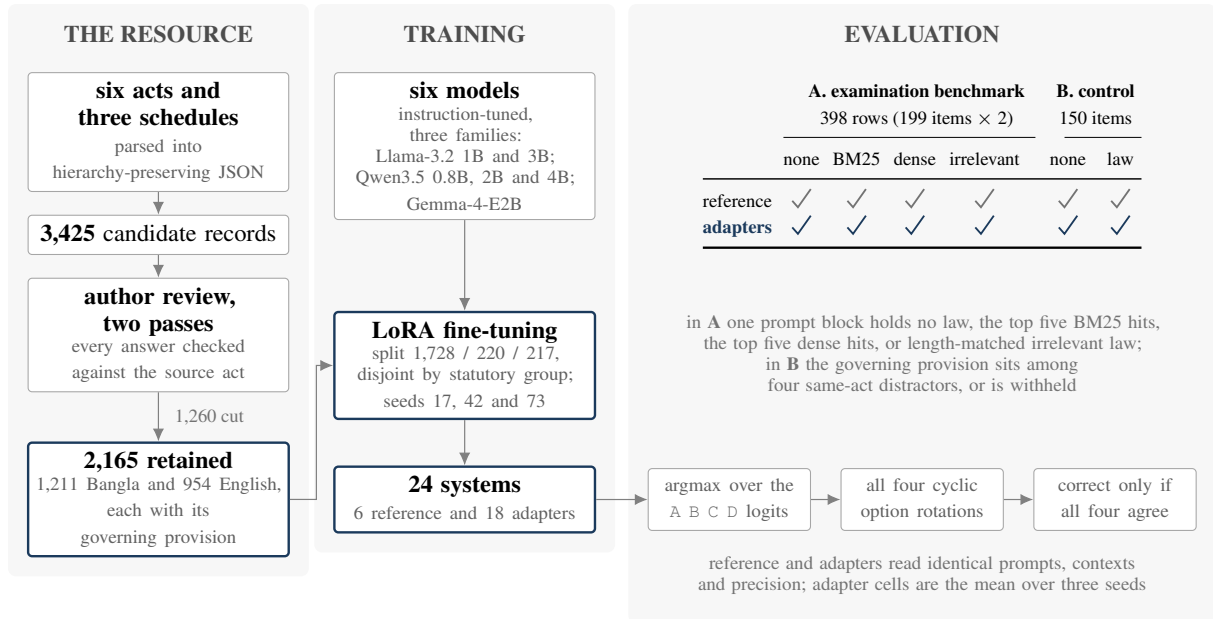

Fine-tuning can appear to improve legal question answering without improving a model's use of the law provided in context. In multiple-choice legal QA, observed accuracy depends not only on whether a model identifies the correct answer, but also on whether the relevant law was retrieved, whether the model used that law correctly, and whether the evaluation procedure correctly interprets the model's output. Conflating these sources of error can therefore produce misleading conclusions about what fine-tuning has actually learned. We study MCQs where sections of Bangladeshi law are supplied
as context. We observe that a wrong answer can reflect three different failures: retrieval
may miss the relevant section, the model may fail to use a section that is
present, or the evaluation procedure may misread the model's answer.
Figure~\ref{fig:design} summarizes how our evaluation separates these
failures.

We evaluate six instruction-tuned models before and after LoRA
fine-tuning. For each model, we train three adapters on 2,165 reviewed
bilingual examples. Each training example pairs a question with the
section of Bangladeshi law needed to answer it. We evaluate on the
2022--2023 Bangladesh Bar Council examinations in Bangla and
machine-translated English.

Fine-tuning also teaches the required MCQ answer format,
\texttt{FINAL ANSWER: (A|B|C|D)}. This creates a scoring problem:
an exact-line parser can reward correct formatting as well as the answer
itself. On the same 398 Bar Council rows, the Qwen3.5-2B seed-42 adapter
gains 50.0\% over the original model under exact-line parsing, but
only 3.0\% when the answer is read from next-token scores. For Gemma-4-E2B,
the two methods favor different systems.

Our primary evaluation therefore avoids parsing generated answers. We
append \texttt{FINAL ANSWER} and select the highest-scoring answer
letter, following option-ID scoring \citep{robinson2023leveraging}.
Option order is a separate source of instability. We therefore rotate the
four choices through all four positions and require the correct answer in
every rotation, following CircularEval \citep{liu2024mmbench}.

In this way, the error in parsing is reduced, but retrieval remains uncertain. As Bar Council
keys do not identify the governing section, a wrong answer cannot indicate
whether the retrieval missed the right section or the model failed to use it.
We therefore construct 150 new bilingual MCQs. Each contains the governing
section and four distractor sections from the same act, so the right
section is guaranteed to be present. Before fine-tuning, five of six
models gain 14.7--19.3\% over no law under the four-order criterion;
all five 95\% confidence intervals lie above zero. For these five models,
we then replace the governing section with another section from the same
act, leaving five sections in the prompt but no governing section.
Accuracy falls by 8.0--15.3\% before fine-tuning and 13.8--14.9\%
after fine-tuning.


To test whether fine-tuning specifically increases reliance on supplied law, we compare the benefit of the governing section before and after fine-tuning. Difference-in-differences estimates show no detectable increase in reliance on the governing section.

\noindent\textbf{Contributions:}
We contribute (1) a hierarchy-preserving Bangladesh statutory corpus and an author-reviewed bilingual fine-tuning set; (2) a 150-item supplied-law control and an evaluation protocol that separates scoring, retrieval, and model failure; and (3) a six-model, three-seed evaluation showing that fine-tuning changes answer behavior but does not detectably increase reliance on the supplied governing law. We release per-item outputs, frozen retrieval inputs, manifests, and analysis scripts to support reproducibility (Appendix~\ref{app:repro}).

\section{Related Work}

\textbf{Bangladesh legal QA.}
MINA evaluates a bilingual, tool-using assistant on the same 2022 and
2023 Bar Council examinations \citep{wasi2026mina}. Its strongest MCQ
systems combine a proprietary model with retrieval or tools and average
five runs without reported option-order rotation. A second study evaluates four
models on 250 Bangla questions from a legal-help forum; three licensed legal professionals evaluate the same set and identify fabricated citations and unsafe advice
\citep{aftahee2025assessing}. These studies establish the local legal QA
setting, but neither isolates whether an error comes from missing law,
failure to use supplied law, or scoring.

\textbf{Legal benchmarks and retrieval.}
LexGLUE standardizes seven English legal NLU datasets, including
five-choice CaseHOLD \citep{chalkidis2022lexglue}, while LegalBench covers
162 tasks across six forms of legal reasoning
\citep{guha2023legalbench}. Our setting is narrower: whether small models
use supplied sections of Bangladeshi law to select answers in Bangla and
English. Retrieval studies show why context availability must be
separated from context use. LegalBench-RAG emphasizes precise legal-span retrieval over broad chunks \citep{pipitone2024legalbenchrag};
NitiBench finds that existing retrievers struggle with complex Thai legal queries \citep{akarajaradwong2025nitibench}; and disrupting
coherence in retrieved United States opinions changes recall
non-monotonically \citep{xia2025beyond}. Our supplied-law control instead
guarantees that the governing section is present, allowing context use to
be tested without retrieval failure.

\textbf{MCQ evaluation.}
MCQ results can also depend on evaluation design. Option reordering can
shift accuracy by 13\% to 85\% across option orders
\citep{pezeshkpour2024large}, while option-ID priors can bias selection
\citep{zheng2024large}. Prompt format can reverse model comparisons
\citep{sclar2024quantifying}, and minor evaluation changes can move
leaderboard ranks by up to eight places
\citep{alzahrani2024benchmarks}. Implementation
choices therefore matter for reproducible evaluation
\citep{biderman2024lessons}. In legal evaluation, GreekBarBench validates
model judges against experts \citep{chlapanis2025greekbarbench}, while
\citet{held2025contemporary} report format validity rising from 0.620 to
0.992 after fine-tuning even though another model still wins the task.

We adopt two established controls unchanged. CircularEval requires
correctness under every cyclic option shift \citep{liu2024mmbench}.
Option-ID scoring reads label logits rather than generated text
\citep{robinson2023leveraging}, avoiding surface-form competition that
can misrank choices \citep{holtzman2021surface}. Together with our
supplied-law control, these methods separate scoring and retrieval failure
from model answer selection.

\section{Data and Evaluation Sets}
\label{sec:data}

\textbf{Corpus.}
The corpus covers six acts in the Bangladesh Bar Council syllabus: the
Penal Code 1860, Code of Criminal Procedure 1898, Evidence Act 1872,
Code of Civil Procedure 1908, Specific Relief Act 1877, and Limitation
Act 1908. We also include CPC Schedule I, CrPC Schedule II, and
Limitation Act Schedule I. English text comes from the official
Bangladesh law portal (\url{http://bdlaws.minlaw.gov.bd/}); Bangla text
comes from published act collections. Appendix~\ref{app:corpus} records
the sources and the conversion that preserves the hierarchy from act to
clause.

\textbf{Fine-tuning data.}
We construct three types of training records. A \emph{single-hop} record
starts from one legal section and asks a question that can be answered
directly from that section. An \emph{advanced-selection} record starts
from five sections of the same act: one contains the rule needed to
answer the question and four are distractors. A \emph{bar-exam-style}
record starts from an MCQ drawn from examination materials other than the
2022--2023 examinations used for evaluation. It is paired with the
section needed to answer it and four distractor sections.
Appendix~\ref{app:qagen} gives the construction details.

These procedures produced 3,425 candidate records. Two author review
passes retained 2,165: 1,211 Bangla and 954 English. The retained set
contains 795 single-hop, 712 advanced-selection, and 658 bar-exam-style
records (Appendix~\ref{app:composition}). Every retained answer was checked
against its cited section, and incomplete, repetitive, or cross-act items
were removed. No legal expert reviewed the full fine-tuning set.

\textbf{Training format.}
Although advanced-selection and bar-exam-style records are constructed
using five sections, the final fine-tuning prompt supplies only the one
section needed to answer the question. We call this the governing section.
For MCQs, the prompt requires the model to return one line in the form
\texttt{FINAL ANSWER: (A|B|C|D)} and maps Bangla option labels to these
letters. The 1,728-row training split contains 541 MCQs, so fine-tuning
directly teaches this answer format. The supplied training section has a
median length of 209 characters. At evaluation, retrieved context instead
contains five sections, with a median rendered length of 2,002
characters.

\subsection{Bar Council Evaluation Set}

We evaluate on the original Bangla questions and official answer keys
from the 2022 and 2023 Bangladesh Bar Council examinations, together with
English machine translations. We exclude cancelled item 2022-005 in both
languages. This leaves 199 unique questions. Each question is evaluated
in Bangla and English, giving 398 question--language pairs. Different
retrieval contexts, option rotations, and training seeds repeat these
same questions; they do not create additional independent test questions.

The English translations received no legal-expert review. We therefore
audit them with two model judges from a vendor different from the
translator (Appendix~\ref{app:transaudit}); this is an audit, not legal
validation. To test whether the judges can detect obvious translation
errors, we deliberately introduce 16 clear option substitutions. The
weaker judge detects only 3. The stronger judge also flags 29 otherwise
unchanged pairs (Table~\ref{tab:judges}). Because those flags may include
real translation problems or false positives, we repeat the English
analysis after removing the 8 items it judges answer-changing and, more
conservatively, after removing all 25 flagged items that never received
an artificial error. The estimates change by at most 0.8\% and 2.1\%,
respectively, and no result changes sign
(Appendix~\ref{app:transrobust}). These checks make gross translation
errors unlikely to explain the English results, but they do not establish
legal correctness or replace expert review.

The 2022--2023 examination questions were not used to create the training
records. We also compare every examination question with every
same-language training question and find no exact match. On a
character-level similarity scale from 0 to 1, where larger values mean
more similar wording, the closest pair scores 0.717. A second comparison
based on shared five-character spans reaches 0.463. The closest pair
tests the same Limitation Act rule but uses different wording. These
checks measure wording overlap only; they do not rule out overlap in
legal content or exposure during pretraining.

\subsection{Supplied-Law Control}
\label{sec:control}

The Bar Council answer keys identify the correct option but not the legal
section needed to answer each question. We therefore cannot tell whether
a wrong answer occurs because retrieval failed to provide the right
section or because the model failed to use a section that was present.
To remove this uncertainty, we construct 150 new MCQs, balanced across
Bangla and English and the six acts. Each question is shown with five
sections from the same act: the governing section and four distractors.
The section needed to answer the question is therefore always present.
The full context has a median length of 1,643 characters and a maximum of
1,948, within the 2,000-character evaluation budget.

Because these questions are newly constructed, we check whether simple
shortcuts can reveal the correct answer without actually using the law.
Choosing the longest option gives 26\% accuracy. A second rule based only
on answer length gives 25\%. Choosing the option that shares the most text
with the supplied law gives 37\%. Random guessing gives 25\%. These checks
test whether obvious surface patterns can solve the questions; they are
not our main evaluation, which rotates all four answer choices and
requires the correct answer in every rotation.

GPT-5.6-Luna drafted the control questions in Bangla and English. It is
not from any evaluated model family, and the control was fixed before we
examined outputs from the evaluated models. Two authors then checked every
question, option, answer key, and cited section and removed 20 items.

\textbf{Independent legal review.}
After all experiments, two legally trained reviewers independently assess
every control item against its governing section. Neither sees model
outputs, experimental results, or the other reviewer's ratings. Both
confirm every answer key, and no key is changed
(Table~\ref{tab:expert}).

\begin{table}[t]
\centering
\small
\setlength{\tabcolsep}{4pt}
\begin{tabular}{lccc}
\toprule
Reviewer & Good & Acceptable & Needs work \\
\midrule
Law graduate & 120 & 28 & 2 \\
Law master's student & 148 & $-$ & 2 \\
\bottomrule
\end{tabular}
\caption{Review of the 150 supplied-law control items. The 28 acceptable
ratings cite minor style concerns (option length, distractor difficulty,
terminology mixing). The four items flagged as needing work do not
overlap. Both reviewers confirmed that no answer is incorrect given its
cited provision. No items were changed; per-item ratings are not
released.}
\label{tab:expert}
\end{table}

The release preserves the exact inference contexts. Schedule text remains
English for Bangla schedule items; 14 of 75 Bangla control items contain
at least one majority-English passage. Reference and adapter always
receive the same string.

\section{Method}
\label{sec:method}

The method separates three effects: whether scoring changes the measured
fine-tuning gain, whether models use a legal section known to be present,
and whether fine-tuning increases the benefit of that section.

\textbf{Training and evaluation context.}
Fine-tuning supplies one governing section per question; retrieval
evaluation supplies five candidate sections. Training on five sections
would roughly triple median prompt length and exceed our 1,536-token
training budget on a single T4 GPU. We therefore test transfer from
one-section training to evaluation with four distractors.

\textbf{Scope.}
We measure whether statutory text changes answer selection, not legal
reasoning or practice. Bar Council keys identify the correct option but
not the governing section. BM25 and dense retrieval therefore describe
how sections are retrieved, not a guarantee that the correct section is
present.

\textbf{Direct option scoring.}
The primary scorer avoids generated-text parsing. We append
\texttt{FINAL ANSWER: (} to the prompt and select the highest-scoring
next token among A, B, C, and D. This removes format parsing. We call
this constrained option-letter scoring, following option-ID scoring
\citep{robinson2023leveraging}.

\textbf{Option rotation.}
Models may prefer an answer position independently of the answer text. We
therefore rotate the four choices through all positions while moving the
correct key with its text. An item is correct only if the model selects
the correct answer in every rotation. This is CircularEval
\citep{liu2024mmbench}; we call the outcome strict consistency. Random
guessing scores $(1/4)^4=0.39\%$, and always choosing one letter scores
zero. All main results use strict consistency unless stated otherwise.

\textbf{Model pairs.}
For each model, the original instruction-tuned checkpoint is the
\emph{reference}; the fine-tuned system (FT) adds an unmerged LoRA adapter
\citep{hu2022lora}. We train adapters with seeds 17, 42, and 73. Within
each pair, model revision, template, prompt, context, precision, runtime,
and scorer are identical.

We evaluate Llama-3.2 at 1B and 3B \citep{meta2024llama}; Qwen3.5 at
0.8B, 2B, and 4B \citep{qwen2026}; and Gemma-4-E2B, with 2.3B effective
and 5.1B total parameters \citep{gemma2026}. Qwen uses FP16 LoRA; Llama
and Gemma use NF4 QLoRA \citep{dettmers2023qlora}.

\textbf{Evaluation conditions.}
The Bar Council evaluation uses no legal context (\textbf{none}), the top
five sections from lexical BM25 retrieval
\citep{robertson2009probabilistic}, or the top five from a fixed BGE-M3
dense retriever \citep{chen2024m3}. Retrieved context is capped at 2,000
characters. The supplied-law control gives five same-act sections,
including the governing section. Its matched \textbf{withheld} condition
replaces that section with another from the same act, leaving five
sections but no governing section. A length-matched \textbf{irrelevant}
condition tests whether additional legal text alone changes performance.
Appendix~\ref{app:chronology} records the protocol chronology and repair
of this condition.

\textbf{Scorer diagnostic.}
To isolate the effect of scoring, we also generate free-form answers for
the same 398 Bar Council rows under dense retrieval and one fixed option
order. For each reference and seed-42 adapter, we compare an exact parser
that accepts only \texttt{FINAL ANSWER: ([ABCD])}, a lenient parser that
searches the output for an answer letter, and the direct scorer above.

\section{Experimental Setup}
\label{sec:setup}

\textbf{Data split.}
We divide the 2,165 fine-tuning records into training, validation, and
internal test sets of 1,728, 220, and 217 records. Records based on the
same act and section stay in the same split. The 2022--2023 Bar Council
benchmark is kept separate and is not used during training or model
selection. Appendix~\ref{app:training} gives the full training details.

\textbf{Statistical analysis.}
The 199 Bar Council questions are the independent test items; languages,
contexts, option orders, and training seeds repeat questions rather
than add independent items. Tables report each original model and mean
of three fine-tuned adapters. All accuracy differences are absolute on the
0--100 percentage scale, not relative changes.

Confidence intervals use 10,000 question-grouped bootstrap samples. We use
two-sided exact McNemar tests for paired strict outcomes, with Holm correction
across 12 tests per model: two languages, two retrieval methods, and three
fine-tuning seeds.

\textbf{Does fine-tuning increase the benefit of supplied law?}
A fine-tuned model can improve even when no law is provided. A general
fine-tuning gain therefore does not show that the model has become more
dependent on the supplied law. We instead compare how much context helps
before and after fine-tuning.

Let $\mathrm{Ref}_C$ and $\mathrm{FT}_C$ denote strict-consistency
accuracy before and after fine-tuning when context $C$ is supplied:
\[
(\mathrm{FT}_C-\mathrm{FT}_{\text{none}})
-
(\mathrm{Ref}_C-\mathrm{Ref}_{\text{none}}).
\]

The first difference measures how much context helps after fine-tuning.
The second measures how much it helps before fine-tuning. A positive
value means the context provides a larger benefit after fine-tuning. We
call this \emph{context specificity}.

For the supplied-law control, $C$ is the condition containing the
governing section. For the Bar Council evaluation, the main table
averages the BM25 and dense estimates because both retrieval methods are
tested. The pre-specified fallback uses dense retrieval versus no
context. Appendix~\ref{app:did} reports both retrieval methods
separately. No sign changes under the dense-only analysis.

Bootstrap intervals for these descriptive comparisons are unadjusted.
Holm correction applies only to the McNemar tests above.

\begin{table*}[t]
\centering
\small
\setlength{\tabcolsep}{2.5pt}
\begin{tabular}{l rrr rrr ccc}
\toprule
 & \multicolumn{3}{c}{reference (\%)} & \multicolumn{3}{c}{FT mean (\%)}
 & \multicolumn{3}{c}{FT $-$ ref (95\% CI)} \\
\cmidrule(lr){2-4}\cmidrule(lr){5-7}\cmidrule(lr){8-10}
Model & none & law & w/o & none & law & w/o & none & law & w/o \\
\midrule
Llama-3.2-1B & 28.7 & 27.3 & 24.0 & \textbf{44.9} & 41.8 & 35.3
 & $+$16.2 [$+$10.4, $+$22.4] & $+$14.4 [$+$7.6, $+$21.3] & $+$10.7 [$+$4.2, $+$17.3] \\
Qwen3.5-0.8B & 38.7 & 57.3 & 49.3 & 52.0 & \textbf{71.8} & 57.8
 & $+$13.3 [$+$7.8, $+$19.3] & $+$14.4 [$+$8.0, $+$21.1] & $+$8.4 [$+$2.0, $+$15.1] \\
Llama-3.2-3B & 50.7 & 68.0 & 58.0 & 53.8 & \textbf{70.2} & 55.3
 & $+$3.1 [\m2.2, $+$8.4] & $+$2.2 [\m4.7, $+$8.9] & \m2.7 [\m9.1, $+$3.6] \\
Qwen3.5-2B & 57.3 & 76.7 & 64.0 & 60.0 & \textbf{78.4} & 63.6
 & $+$2.7 [\m2.0, $+$7.3] & $+$1.8 [\m3.8, $+$7.3] & \m0.4 [\m6.9, $+$6.2] \\
Gemma-4-E2B & 70.0 & \textbf{84.7} & 69.3 & 58.9 & 78.7 & 64.4
 & \m11.1 [\m17.6, \m4.9] & \m6.0 [\m11.3, \m0.7] & \m4.9 [\m9.8, \m0.2] \\
Qwen3.5-4B & 76.7 & \textbf{94.0} & 79.3 & 69.1 & 91.1 & 77.1
 & \m7.6 [\m13.6, \m1.8] & \m2.9 [\m6.9, $+$1.1] & \m2.2 [\m6.9, $+$2.4] \\
\bottomrule
\end{tabular}
\caption{Results on the 150 supplied-law control questions using strict
consistency, which requires the correct answer in all four option orders.
\textbf{none} gives no legal context; \textbf{law} gives the governing
section plus four same-act distractors; \textbf{w/o} uses the same
five-section setup but replaces the governing section with another
same-act section. Ref is the original model and FT mean is the mean over
three fine-tuning seeds. FT$-$ref is their accuracy difference within
each condition; brackets show the 95\% confidence interval. Bold marks
the highest accuracy in each row.}
\label{tab:control}
\end{table*}

\begin{table*}[t]
\centering
\small
\setlength{\tabcolsep}{3pt}
\begin{tabular}{l rr rr rr rr cc c}
\toprule
 & \multicolumn{2}{c}{none} & \multicolumn{2}{c}{BM25}
 & \multicolumn{2}{c}{dense} & \multicolumn{2}{c}{irrelevant}
 & \multicolumn{3}{c}{context specificity} \\
\cmidrule(lr){2-3}\cmidrule(lr){4-5}\cmidrule(lr){6-7}\cmidrule(lr){8-9}\cmidrule(lr){10-12}
Model & Ref & FT & Ref & FT & Ref & FT & Ref & FT
 & R$-$N & R$-$I & Holm \\
\midrule
\multicolumn{12}{l}{\emph{Bangla}} \\
Llama-3.2-1B & 0.5 & \textbf{5.5} & 0.5 & \textbf{8.5} & 2.0 & \textbf{13.1} & 0.5 & \textbf{6.0}
 & $+$4.5 [$+$0.9, $+$8.1] & $+$4.0 [$+$0.0, $+$8.2] & 6/6 \\
Qwen3.5-0.8B & 3.5 & \textbf{4.0} & 11.6 & \textbf{17.3} & 19.6 & \textbf{24.1} & 6.0 & \textbf{7.4}
 & $+$4.6 [$+$1.2, $+$8.0] & $+$3.8 [\m0.4, $+$7.9] & 1/6 \\
Llama-3.2-3B & \textbf{6.5} & 5.7 & 12.1 & \textbf{17.1} & 14.1 & \textbf{23.6} & 4.0 & \textbf{8.4}
 & $+$8.1 [$+$3.7, $+$12.6] & $+$2.9 [\m1.8, $+$7.7] & 3/6 \\
Qwen3.5-2B & 2.5 & \textbf{6.4} & 19.1 & \textbf{20.8} & 23.1 & \textbf{26.8} & 6.5 & \textbf{7.2}
 & \m1.2 [\m5.5, $+$3.0] & $+$2.0 [\m2.2, $+$6.1] & 0/6 \\
Gemma-4-E2B & \textbf{8.5} & 7.5 & \textbf{27.1} & 26.3 & \textbf{37.7} & 32.8 & \textbf{11.1} & 6.4
 & \m1.8 [\m6.2, $+$2.6] & $+$1.8 [\m2.1, $+$5.9] & 0/6 \\
Qwen3.5-4B & 9.5 & \textbf{11.7} & 24.1 & \textbf{27.6} & 28.6 & \textbf{36.5} & \textbf{11.1} & 10.9
 & $+$3.5 [\m0.3, $+$7.2] & $+$5.9 [$+$1.8, $+$10.2] & 2/6 \\
\midrule
\multicolumn{12}{l}{\emph{English}} \\
Llama-3.2-1B & 8.5 & \textbf{15.4} & 11.6 & \textbf{25.5} & 15.6 & \textbf{34.8} & 4.0 & \textbf{10.2}
 & $+$9.7 [$+$4.3, $+$15.2] & $+$10.4 [$+$5.1, $+$15.5] & 6/6 \\
Qwen3.5-0.8B & 11.6 & \textbf{13.2} & 24.6 & \textbf{33.2} & 31.2 & \textbf{41.4} & 7.0 & \textbf{11.2}
 & $+$7.7 [$+$3.1, $+$12.3] & $+$5.2 [$+$0.1, $+$10.2] & 6/6 \\
Llama-3.2-3B & \textbf{20.1} & 19.1 & 34.7 & \textbf{37.5} & \textbf{43.7} & 42.9 & 17.6 & \textbf{18.8}
 & $+$2.0 [\m3.4, $+$7.3] & \m0.2 [\m5.3, $+$4.9] & 0/6 \\
Qwen3.5-2B & 14.6 & \textbf{20.8} & 36.7 & \textbf{38.7} & 38.7 & \textbf{45.4} & 13.6 & \textbf{15.9}
 & \m1.8 [\m6.9, $+$3.3] & $+$2.0 [\m3.2, $+$7.3] & 0/6 \\
Gemma-4-E2B & 21.1 & 21.1 & \textbf{37.7} & 35.8 & \textbf{49.7} & 45.2 & \textbf{19.6} & 15.6
 & \m3.2 [\m7.5, $+$1.3] & $+$0.8 [\m3.6, $+$5.4] & 0/6 \\
Qwen3.5-4B & 28.1 & \textbf{29.6} & \textbf{44.2} & 41.2 & 50.8 & \textbf{53.1} & \textbf{25.1} & 24.0
 & \m1.8 [\m7.0, $+$3.1] & $+$0.8 [\m3.3, $+$5.0] & 0/6 \\
\bottomrule
\end{tabular}
\caption{Results on the 199 Bar Council questions using strict
consistency, which requires the correct answer in all four option orders.
Ref is the original model; FT is the mean over three fine-tuning seeds.
R$-$N measures whether fine-tuning changes the benefit of retrieved law
relative to no legal context. R$-$I makes the same comparison using
length-matched irrelevant legal text instead of no context. Holm reports
how many of the six McNemar tests remain significant after correction.
Bold marks the higher Ref or FT score within each condition, not
statistical significance.}
\label{tab:exam}
\end{table*}

\section{Results}
\label{sec:results}

Results follow the evaluation sequence in
Figure~\ref{fig:design}. We first show how scoring changes the
measured fine-tuning gain, then test whether models use a
governing section known to be present. Finally, we return to
Bar Council questions, where that section must be retrieved.

\subsection{Scoring Can Change the Measured Fine-Tuning Gain}
\label{sec:route}

We score the same 398 dense-retrieval rows at one fixed answer
order in three ways. For Qwen3.5-2B, reference accuracy is
4.0\% with exact-line parsing, 33.2\% with lenient parsing, and
51.0\% with direct A/B/C/D scoring. The seed-42 adapter scores
54.0\% under all three. The measured fine-tuning gain is
therefore $+$50.0\%, $+$20.9\%, or $+$3.0\% on the same model
and questions. Tables~\ref{tab:scorer} and~\ref{tab:fourways} show all six models.

\begin{table}[!t]
\centering
\small
\setlength{\tabcolsep}{4pt}
\begin{tabular}{lrrrr}
\toprule
 & format & \multicolumn{3}{c}{FT gain} \\
\cmidrule(lr){2-2}\cmidrule(lr){3-5}
Model & Ref.\ (\%) & exact-line & lenient & logit \\
\midrule
Qwen3.5-2B    &  5.8 & $+$50.0 & $+$20.9 & $+$3.0 \\
Llama-3.2-1B  &  6.0 & $+$39.7 & $+$10.3 & $+$1.8 \\
Qwen3.5-0.8B  & 80.7 &  $+$9.8 &  $+$6.5 & $+$3.5 \\
Gemma-4-E2B   & 80.9 & $+$10.1 & $+$10.1 & \m2.5 \\
Llama-3.2-3B  & 81.7 &  $+$9.3 &  $+$0.8 & $+$2.8 \\
Qwen3.5-4B    & 98.7 &   \m0.8 &   \m1.3 & $+$0.0 \\
\bottomrule
\end{tabular}
\caption{Effect of scoring method on the measured fine-tuning gain for
the same 398 dense-retrieval rows. Ref.\ format is the percentage of
reference outputs that follow the required answer format. FT gain is
seed-42 adapter accuracy minus reference accuracy.}
\label{tab:scorer}
\end{table}

The gap comes partly from formatting. Fine-tuned MCQ models were
trained to produce the required answer line; reference models
were not. Qwen3.5-2B follows that format in only 5.8\% of
reference generations. Exact-line parsing therefore measures
formatting as well as answer selection; direct A/B/C/D scoring
removes that step.

Scoring can also change which system wins. For Gemma-4-E2B,
exact-line parsing gives the adapter a $+$10.1\% advantage,
while direct scoring gives a \m2.5\% disadvantage. This reversal holds across seeds (Appendix~\ref{app:addl}), although the
Gemma adapter shows signs of over-adaptation. For Qwen3.5-4B, format compliance is 98.7\%,
and the two gains differ by only 0.8\%. The scorer effect is
therefore not monotonic across models.

Answer order creates a separate instability. Under dense
retrieval, reference-model Bangla position spread ranges from
13.6\% to 57.3\% (Table~\ref{tab:posbias}).
Table~\ref{tab:rotation} shows the drop when correctness is
required in all four orders.

\begin{table}[!t]
\centering
\small
\setlength{\tabcolsep}{3.5pt}
\begin{tabular}{lrrrrrr}
\toprule
 & \multicolumn{3}{c}{Bangla} & \multicolumn{3}{c}{English} \\
\cmidrule(lr){2-4}\cmidrule(lr){5-7}
System & rot0 & strict & drop & rot0 & strict & drop \\
\midrule
Llama-3.2-1B  & 34.7 &  2.0 & 32.7 & 47.2 & 15.6 & 31.7 \\
Llama-3.2-3B  & 42.7 & 14.1 & 28.6 & 61.3 & 43.7 & 17.6 \\
Qwen3.5-0.8B  & 44.2 & 19.6 & 24.6 & 51.8 & 31.2 & 20.6 \\
Qwen3.5-2B    & 45.2 & 23.1 & 22.1 & 56.8 & 38.7 & 18.1 \\
Qwen3.5-4B    & 55.8 & 28.6 & 27.1 & 67.8 & 50.8 & 17.1 \\
Gemma-4-E2B   & 54.8 & 37.7 & 17.1 & 61.8 & 49.7 & 12.1 \\
\bottomrule
\end{tabular}
\caption{Effect of option order on reference models under dense
retrieval. rot0 is accuracy at one fixed order; strict requires the
correct answer in all four orders; drop is rot0 minus strict.}
\label{tab:rotation}
\end{table}

All results below therefore use direct A/B/C/D scoring and
strict consistency.

\subsection{Supplied Law Helps Five of Six Models}
\label{sec:lawhelps}

We next remove retrieval uncertainty by guaranteeing that the
governing section is present. On the same 150 control questions,
Table~\ref{tab:control} compares no law with five sections that
include the governing section. Five of six reference models gain
14.7--19.3\% in strict accuracy. All five confidence intervals
exclude zero, with lower bounds of at least $+$6.7\%
(Table~\ref{tab:intervals}). Qwen3.5-4B reaches 94.0\%.

A post-hoc check tests whether lexical overlap explains the gain.
On 92 questions where a wrong option shares more five-character
spans with the law than the correct option, the same five models
still gain 12.0--22.8\% (Appendix~\ref{app:addl}).
Llama-3.2-1B remains the exception at \m2.2\%.

Llama-3.2-1B also shows why rotation matters. Its strict accuracy
moves from 28.7\% without law to 27.3\% with law, a \m1.3\%
change with interval [\m7.3\%, $+$4.7\%]. At one fixed order,
however, supplied law raises accuracy by 6.7\%. Llama-3.2-3B,
from the same family and release, gains 17.3\% under strict
consistency.

To test whether the governing section itself drives the gain, we
remove only that section while keeping the four distractors.
For the five responsive reference models, accuracy falls by
8.0--15.3\%, with every confidence interval excluding zero
(Table~\ref{tab:withheld}). Their three-seed adapter means fall
by 13.8--14.9\%. These models therefore depend on the governing
section, not merely on other text from the same act.

\subsection{Strict Consistency Measures Correctness and Position Stability}
\label{sec:rotdecomp}

Strict consistency can improve because supplied law raises
fixed-order accuracy, improves stability when options move, or
both. Table~\ref{tab:rotdecomp} separates these effects.

\begin{table}[!t]
\centering
\small
\setlength{\tabcolsep}{3.5pt}
\begin{tabular}{lrrrrrr}
\toprule
 & \multicolumn{2}{c}{law} & \multicolumn{2}{c}{no law}
 & \multicolumn{2}{c}{law effect} \\
\cmidrule(lr){2-3}\cmidrule(lr){4-5}\cmidrule(lr){6-7}
Reference & rot0 & strict & rot0 & strict & rot0 & strict \\
\midrule
Llama-3.2-1B & 63.3 & 27.3 & 56.7 & 28.7 & $+$6.7 & \m1.3 \\
Qwen3.5-0.8B & 74.7 & 57.3 & 58.0 & 38.7 & $+$16.7 & $+$18.7 \\
Llama-3.2-3B & 84.7 & 68.0 & 73.3 & 50.7 & $+$11.3 & $+$17.3 \\
Qwen3.5-2B   & 88.0 & 76.7 & 76.7 & 57.3 & $+$11.3 & $+$19.3 \\
Gemma-4-E2B  & 92.7 & 84.7 & 79.3 & 70.0 & $+$13.3 & $+$14.7 \\
Qwen3.5-4B   & 98.7 & 94.0 & 86.7 & 76.7 & $+$12.0 & $+$17.3 \\
\bottomrule
\end{tabular}
\caption{Reference-model accuracy on the 150 supplied-law control
questions. ``law'' means the governing section is included with four
same-act distractors; ``no law'' means no legal context is supplied.
rot0 is accuracy at one fixed answer order, while strict requires the
model to answer correctly in all four answer orders. Law effect is the
difference between the law and no-law accuracies.}

\label{tab:rotdecomp}
\end{table}

At one fixed order, supplied law improves accuracy for all six
models. For the five that also improve under strict consistency,
the law effect is 1.3--8.0\% larger under strict scoring.
This suggests that supplied law also reduces answer-position sensitivity.
Because fixed-order accuracy rises too, stability alone cannot
explain the strict gain.

Llama-3.2-1B differs: its $+$6.7\% fixed-order effect becomes
\m1.3\% under strict consistency. Supplied law helps at one
order, but the gain does not survive when the choices move.

\subsection{Fine-Tuning Does Not Detectably Increase Dependence on Law}

We next ask whether fine-tuning increases the benefit of supplied
law. Without law, the two reference models below 40\% gain
13.3\% and 16.2\% after fine-tuning. The next two gain 2.7\%
and 3.1\%, while the two at or above 70\% lose 7.6\% and
11.1\%. With only six models, this is descriptive, not a scaling
or model-family law.

Higher accuracy after fine-tuning does not show that supplied law
became more useful. On the control, difference-in-differences
estimates range from \m1.8\% to $+$5.1\%, and every confidence
interval includes zero (Table~\ref{tab:intervals}). For the four
reference models below 60\%, fine-tuning effects with and without
law differ by at most 1.8\%; the two strongest models decline in
both conditions. Fine-tuning changes answer selection and
preserves dependence on the governing section, but we detect no
increase in that dependence.

\subsection{Bangla Scores Lower, but the Law Effect Persists}

Bar Council accuracy is lower in Bangla than in English
(Table~\ref{tab:exam}). On the supplied-law control, however,
the five responsive models gain 13.3--21.3\% in Bangla and
12.0--21.3\% in English. Llama-3.2-1B remains the exception at
\m4.0\% and $+$1.3\%. Removing translation pairs flagged in the
audit leaves every context-specific sign unchanged
(Appendix~\ref{app:transrobust}).

\subsection{The Bar Council Set Gives Weaker Evidence About Context Use}

We finally return to the Bar Council questions. Their answer
keys do not identify the governing section, so retrieval failure
and failure to use retrieved law cannot be separated as in the
supplied-law control.

Three models show positive difference-in-differences estimates with confidence intervals above zero 
when retrieval is compared with no context: $+$7.1\% for
Llama-3.2-1B, $+$6.2\% for Qwen3.5-0.8B, and $+$5.1\% for
Llama-3.2-3B. We call this comparison R$-$N. Five of its twelve
descriptive confidence intervals exclude zero, all positively.

As a sensitivity check, R$-$I compares retrieval with the
length-matched irrelevant condition instead of no context.
Three R$-$I intervals exclude zero. Nine of the 96 examination
cells are at or below 5\% (Table~\ref{tab:sparse}), limiting
precision. The task also differs from training: adapters see one
clean section during fine-tuning but five candidate sections at
evaluation. Unlike the supplied-law control, the examination
cannot tell whether retrieval supplied the governing section.

\section{Discussion}

The apparent fine-tuning effect changes once scorer, retriever, and
model failure are separated.

Scoring sets the first boundary. Because MCQ training teaches the
required answer format, generate-then-parse evaluation can credit
formatting as legal adaptation. Constrained option scoring removes this
parse failure, while full rotation exposes gains that depend on answer
position. These controls measure answer selection, not explanation
quality \citep{chlapanis2025greekbarbench}.

The supplied-law control removes retrieval uncertainty. Five reference
models improve when the governing provision is present; those models
and their adapters decline when it is removed. Retrieval failure
therefore cannot explain these effects. Llama-3.2-1B remains the
exception: its fixed-order gain disappears under rotation, so position
stability is part of the result.

With scoring and retrieval controlled, fine-tuning changes answer
selection and preserves dependence on the governing provision, but we
detect no increase in that dependence. Family, size, release, and
precision vary across six models, limiting broader conclusions. The
transferable result is the measurement sequence: separate scorer,
retriever, and model failure before interpreting legal adaptation.

\section{Conclusion}

Five of six instruction-tuned checkpoints use supplied Bangladeshi
statute under a rotation-invariant control: the governing provision
improves accuracy by 14.7--19.3\%, and removing it reduces accuracy by
8.0--15.3\% for the five responsive reference models and 13.8--14.9\%
for their adapters. Yet exact-line parsing reports $+$50.0\% where
matched constrained scoring gives $+$3.0\%, and reverses one model
comparison. Single-provision fine-tuning changes answers without a
detected increase in provision dependence; five-candidate training
remains untested. Legal adaptation claims require separating scoring method,
retriever, and model failure.

\section*{Limitations}

\textbf{The task and sample bound the claim.}
The benchmark contains 199 questions from two examination years in one
jurisdiction. Bangla and English rows, contexts, rotations, and seeds
repeat those questions rather than adding independent test items. MCQ
selection is a proxy for statutory knowledge, not legal reasoning or
practice. Strict consistency also reflects option-order stability.

\textbf{Training and evaluation differ by design.}
Training supplies one governing provision, whereas evaluation supplies
five candidate sections. We therefore test transfer from clean law to
distractor-rich prompts, not training directly on that setting.
Five-candidate training remains untested because of the available
single-T4 compute budget. Our null result does not extend to that
intervention.

\textbf{The six checkpoints do not form a scaling study.}
The models differ in family, size, release, and precision, and all
selected the upper edge of the learning-rate grid. Precision is matched
within reference-adapter pairs but not across families. Relations between
starting accuracy and fine-tuning effects are therefore descriptive.

\textbf{Contamination cannot be ruled out.}
Public pretraining may contain the Bar Council examinations. One retained
training item also tests the legal rule underlying one benchmark item,
although the wording is not an exact match. Repeated development on the
same 199 questions creates an additional risk of co-adaptation.

\textbf{Resource validation remains incomplete.}
The English examination translations received no legal-expert review;
model-based audits do not replace expert validation. Authors checked all
2,165 fine-tuning answers against their cited provisions, but no lawyer
exhaustively reviewed the set.

GPT-5.6-Luna drafted the 150 supplied-law control items. Authors checked
every item and removed 20; two legally trained reviewers later confirmed
every retained answer key. The reviewers validated rather than authored
the items.

Schedules exist only in English in our corpus, so 14 of 75 Bangla
control items contain a majority-English passage. Matched systems receive
identical contexts, but this limits language-specific interpretation.

The Bar Council examination provides no expert governing-provision
labels. We therefore cannot report oracle retrieval or Recall@k, and
cannot separate retrieval failure from context-use failure on those
questions. The supplied-law control removes this uncertainty only for
its 150 items.

\textbf{The evaluation-method findings have different limitations.}
Parser inflation follows directly from evaluation that rewards an answer
format taught during fine-tuning. The Gemma-4-E2B sign reversal is less
general because evaluation-method effects and over-adaptation cannot be
separated for that model. A format example at inference time was not tested.

\section*{Ethics Statement}

Legal question answering is sensitive. The evaluated systems are
research artifacts and are not intended to provide legal advice.
Even the strongest configurations fail under strict scoring, and the
task omits drafting, citation verification, client facts, procedural
judgment, and professional responsibility. Bar-exam performance is not
evidence of competence to advise a person.

The release contains public statutes and examinations and no client
records or personal data. We license generated questions, answers,
annotations, translations, metadata, and arrangement under CC BY 4.0
only where we hold the rights. We claim no ownership over official
statutory text or examination questions. The anonymous supplement
removes author names, personal links, and identifying metadata.

Evaluation design also carries practical risk. A parser that rewards
format compliance can overstate legal-QA progress. Separating answer
selection from formatting and model failure from retrieval failure
limits what can reasonably be inferred from the reported results.

\paragraph{Use of AI Assistants.} The authors used AI writing tools,
including ChatGPT and Claude, to improve manuscript fluency and clarity.
Human authors remain responsible for the ideas, experiments, analyses,
claims, integrity, and accuracy of the paper.

\bibliography{references}

\appendix

\section*{Appendix}

Appendices~\ref{app:corpus}--\ref{app:chronology} document the corpus,
data, training, and protocol chronology;
Appendices~\ref{app:did}--\ref{app:repro} give the retriever
decomposition, the uncertainty intervals behind every reported
difference, the translation audit, and the release.

\section{Corpus Construction}
\label{app:corpus}

\textbf{The corpus follows the Bar Council syllabus.} English statutory
text comes from the official Bangladesh law portal maintained by the
Ministry of Law, Justice and Parliamentary Affairs
(\url{http://bdlaws.minlaw.gov.bd/}). Bangla text comes from published
act collections. The same Bangla renderings recur across several
editions, so we treat them as standard published text without assigning
them to one publisher. The corpus covers the Penal Code 1860 (Act XLV of
1860), Code of Criminal Procedure 1898 (Act V of 1898), Evidence Act 1872
(Act I of 1872), Code of Civil Procedure 1908 (Act V of 1908), Specific
Relief Act 1877 (Act I of 1877), and Limitation Act 1908 (Act IX of
1908), plus CPC Schedule I, CrPC Schedule II, and Limitation Act Schedule
I.

The three schedules are available only in English in our corpus. A Bangla
question governed by a schedule keeps its Bangla question and options but
receives English schedule text. The language label describes the
question, not every passage in its prompt.

\textbf{Parenthesized labels create the parsing problem.} Bangladeshi
statutes nest acts, parts, chapters, sections, subsections, clauses, and
subclauses. Sections use Arabic numerals and a full stop
(\texttt{1.}); subsections use parenthesized Arabic numerals
(\texttt{(1)}); clauses use parenthesized lowercase letters
(\texttt{(a)}); and subclauses use parenthesized Roman numerals
(\texttt{(i)}). Illustrations and explanations may attach at any level.
Illustrations also reuse labels such as \texttt{(a)}, making them
indistinguishable from clauses without context. An explicit heading
always introduces an illustration sequence, so a regular-expression pass
can separate the two before model processing. This pass lets the JSON
retain each illustration under the provision it explains instead of
flattening it into the clause list.

\textbf{Conversion preserves the provision hierarchy.} Each act receives
its own schema-enforcing conversion prompt because formatting conventions
differ across the corpus. The schema creates a root object for the act
name, enactment date, preamble, and parts or chapters. A section stores
its number, marginal note, text, and optional subsections, clauses, and
illustrations; subsections and clauses recurse in the same form.
Act-specific rules remove page headers, running titles, and line numbers;
join lines broken inside sentences; recover marginal notes around section
numbers; and record omitted or repealed sections. We revised each prompt
until its output conformed, then inspected the files by hand.

Section 81 of the Penal Code shows what this buys. The converted object
retains the chapter (\emph{general exceptions}), marginal note, section
text, explanation, and both illustrations as distinct labeled objects. At
inference, the system can supply the provision without turning
illustration numbering into clause noise.

\section{QA Generation and Review}
\label{app:qagen}

\textbf{Three tasks increase the selection difficulty.} Each language has
one prompt for each of three categories, producing six prompt
combinations. Single-hop extraction supplies one provision containing the
answer. Advanced selection supplies exactly five verbatim corpus
provisions, one governing provision and four distractors, and requires
elimination. Bar-exam generation adds examination style. Its reference
questions come from papers other than the 2022 and 2023 benchmark
examinations; the generated item is paired with one governing provision
and four corpus distractors. A \texttt{Cancelled} flag retains voided
reference questions as explicit records. Unlike the official benchmark
keys, bar-exam training keys are created during generation and checked
during review.

\textbf{The governing provision moves, but not uniformly.} Both selection
prompts require the correct candidate to vary from the first through
fifth position with approximately equal frequency. Across 712
advanced-selection records, it appears in position 1 for 13.8\%, position
2 for 50.4\%, position 3 for 23.7\%, position 4 for 7.0\%, and position 5
for 5.1\%. The correct provision does not stay in one slot, but the
distribution has an unintended position bias. Because reference models
also carry positional priors (Section~\ref{sec:results}), the primary
metric scores every benchmark item under all four option orders.

\textbf{The bar-exam category carries the main leakage risk.} Its prompt
sees examination-style reference material, but never the 2022 or 2023
benchmark papers. Reference items provide style only and do not enter the
released records. Every retained record then passes two author checks
against the source law. No same-language bar-exam training and benchmark
pair matches exactly. The maximum character 5-gram Jaccard score is
0.463. \texttt{SequenceMatcher} reaches 0.717 for a Limitation Act
schedule question paired with benchmark item 2022-086. Their wording
differs, but both test the same limitation period. We therefore report
substantive overlap instead of treating the absence of a verbatim match
as proof of separation.

\textbf{Prompts counter five recurring generation failures.}
Verbatim-copy rules prevent statutory paraphrase. Clause enumeration
prevents partial coverage of long sections. A fixed question-type
vocabulary and a no-repeat rule reduce duplicate scenarios. Requiring the
act name first and the citation last guards against cross-act records. A
pre-output checklist counters format drift over long runs. These
constraints act during generation instead of filtering the failures
afterward.

\textbf{Human review retains 2,165 records.} Authors checked every
candidate against the source law. They verified each answer against its
cited provision, checked that statutory quotations were copied without
summary, and deleted or regenerated repetitive, cross-act, or materially
incomplete records. The earliest preserved pre-filter pool contains 3,425
records; 2,165 remain, so 1,260 were removed. Review occurred during
generation, so no per-record log exists. Identifier and question-text
checks confirm that the release is a strict subset
(\texttt{prefilter\_pool.py}).

\paragraph{Generation prompt, single-hop extraction, English.}
Reproduced in the form used, with the example entry abbreviated. The
other five prompts follow the same skeleton with category-specific
constraint blocks, and all six are in the release.

\begin{lstlisting}
Role: Expert Legal Educator and Subject Expert.
Task: Analyze the provided JSON legal dataset and transform each section of the Act into a structured assessment and reference entry.
[SFT Data Quality Context]
You are generating supervised fine-tuning (SFT) data for smaller language models (0.6B-8B parameters). Your outputs are the training signal. Smaller models learn by pattern recognition: they need structural consistency. Length variance and format drift degrade SFT quality directly.
[Jurisdiction & Corpus]
Legal Corpus: Use only the Act names, section numbers, and legislative addresses provided in the dataset. Never invent a Section number if the provision uses a different label.
[Format & Schema]
Output must be a strictly valid JSON array of objects. Schema per object: Section Number; Entry ID; Question (plain layperson language, no legal jargon in the stem); Subsection/Clause; Section Text (full, verbatim); Answer; Type; Difficulty; Keywords; Cited Acts and Sections (must be the last key).
[Field Rules: The Answer Field]
Strict Opening: Must begin with "Under [Act Name]," followed immediately by the precise legislative address. Constraint: one to two sentences maximum, 25-45 words.
[Constraint Guidelines]
Verbatim Integrity: Copy "Section Text" exactly. Do not truncate, paraphrase, or reconstruct from memory. Client-Speak Rule: questions must sound like a client describing a problem. Scenario Freshness: each scenario must use a distinct factual context; do not reuse character names across consecutive entries. Multi-Clause Sections: generate one question targeting the most significant clause and note it in the "Subsection/Clause" field.
[Pre-Output Quality Check]
[ ] Is Section Text copied verbatim? [ ] Does the Answer start with "Under [Act Name]," and stay under 45 words? [ ] Is the Question written in plain, non-legal language? [ ] Is the JSON strictly valid with "Section Number" first and "Cited Acts" last?
\end{lstlisting}

The Bangla prompts are the same skeleton with the instruction text and
the answer-field rules in Bangla, and with an additional type-selection
procedure that draws from a fixed question-type vocabulary and forbids
repeating a type consecutively. They are released as files rather than
reproduced here because this template cannot typeset Bangla script.

\paragraph{Generation prompt, advanced selection, English.} The
constraint block that distinguishes this category from single-hop
extraction. The rest of the prompt shares the skeleton above.

\begin{lstlisting}
Possible Sections: A list of exactly 5 objects, each containing Section Number and Full Text (the correct section + 4 distractors). Full text must be copied verbatim from the law dataset.
The governing section's position - first, second, third, fourth, or fifth - must vary randomly across entries, with approximately equal frequency. Never default to always placing it first.
Multi-Clause Sections: Target the specific clause that governs the scenario and note it in the Subsection/Clause field.
\end{lstlisting}

\paragraph{Generation prompt, bar-exam category, English.} This category
is pinned to a supplied reference file rather than composing freely,
which is why its prompt speaks of copying rather than writing. The
reference collection holds bar-examination-style questions from
examinations other than the two we evaluate on; the model's work here is
locating the governing provision and the four distractors, not inventing
the stem. We reproduce the prompt as run, and note that we can vouch for
what was supplied to it, not for how that reference collection was itself
assembled.

\begin{lstlisting}
Task: You are provided bar examination questions from the supplied JSON dataset, each with multiple-choice options (A, B, C, D) and a known correct answer. For each question, identify the one governing legal provision from the supplied law dataset, produce a full Answer citing the relevant section, and state the correct option letter as the final element of the Answer field.
Question: The verbatim question text from the bar exam JSON, exactly as written. Do not rephrase or correct.
Bar Exam Options: The options object from the bar exam JSON - keys A, B, C, D with their verbatim text. Copy exactly.
Cancelled: Boolean. Set to true if correct answer in the bar exam JSON is "Cancelled" or equivalent. When true, set Relevant Section to "N/A" and Answer to "This question was cancelled in the original exam."
Legal Corpus: Use only the Act names, section numbers, and legislative addresses that appear in the provided law dataset. Do not import terminology or citation formats from other jurisdictions.
\end{lstlisting}

The jurisdiction constraint in the last block exists because closed-book
prompting produced confident citations to Indian statute in place of the
Bangladeshi equivalent, which is the failure the corpus and the supplied
context are meant to remove.

\section{Training Details}
\label{app:training}

All models use LoRA rank and alpha 16 on each linear language projection,
effective batch 16, at most two epochs, and validation-loss early
stopping. A one-epoch 512-record pilot selects from
$\{5\mathrm{e}{-}5, 1\mathrm{e}{-}4, 2\mathrm{e}{-}4\}$. Every model
selects $2\mathrm{e}{-}4$, the upper boundary and Unsloth default
\citep{han2023unsloth}; we do not widen the grid. Each seed trains on one
16\,GB T4.

\paragraph{Frozen prompt.} Every system, in both arms and in training,
receives the same system message:

\begin{lstlisting}
You answer Bangladesh law questions using the supplied legal context when it is relevant. If the question contains answer options, return exactly one line and nothing else in this form: FINAL ANSWER: (A|B|C|D). For Bangla option labels, A, B, C, and D map to the four Bangla option glyphs. If the question has no answer options, answer the question directly.
\end{lstlisting}

The released \texttt{bdlex\_core.py} uses the Bangla option glyphs
verbatim and normalizes Latin, numeric, and Bangla labels. Each user turn
places the context after \texttt{LEGAL CONTEXT:} and the question after
\texttt{QUESTION:}. The none condition instead supplies the literal
string \texttt{[NO CONTEXT PROVIDED]}. Before scoring, marker matching in
both languages removes cancelled items.

For every MCQ, the conversion writes the complete assistant target
\texttt{FINAL ANSWER: (X)}; direct-QA rows retain prose targets. The
executed training split contains 541 MCQs and 1,187 direct-QA rows. At
constrained evaluation, the prompt instead ends at the partial stem
\texttt{FINAL ANSWER: (} because the next-token logits over A, B, C, and
D are the score. No closing parenthesis is generated or parsed on this
route.

\paragraph{Training recipe.} We hold one recipe fixed across all six
models. It uses sequence length 1,536, per-device batch 4, gradient
accumulation 4, and effective batch 16. LoRA uses rank 16, alpha 16, no
dropout, and all linear language-model projections
(\texttt{q,k,v,o,gate,up,down}). Optimization uses 8-bit AdamW, weight
decay 0.01, a cosine schedule, warmup ratio 0.06, and gradient clipping
at 1.0. Gradient checkpointing is on, packing is off, and early stopping
allows one validation-loss evaluation without improvement across at most
two epochs. Because we add no tokens, embeddings and the output head
remain frozen. Training uses Unsloth \citep{han2023unsloth}. Two epochs
at effective batch 16 produce 216 optimizer steps.

We split records by normalized (act, section) group with a fixed splitter
seed. The result is 1,728 training, 220 validation, and 217 internal-test
records, with no statutory group crossing a split. Each model is trained
on one 16\,GB T4 with seeds 17, 42, and 73.

Early stopping fired for Qwen3.5-2B seed 42 and for all three Qwen3.5-4B
seeds, each halting at step 162 with the best checkpoint retained; every
other run completed 216 steps. Final training and validation loss were
0.556/0.568 for Llama-3.2-1B, 0.432/0.460 for Llama-3.2-3B, 0.931/0.903
for Qwen3.5-0.8B, 0.796/0.768 for Qwen3.5-2B, 0.691/0.647 for
Qwen3.5-4B, and 0.132/0.875 for Gemma-4-E2B, the only model whose
validation loss finished far above its training loss.

\paragraph{Learning-rate pilots.} Each pilot runs one epoch with seed 17
on the same deterministic 512-row training subset. Selection uses grouped
validation loss over the pre-specified grid
$\{5\mathrm{e}{-}5, 1\mathrm{e}{-}4, 2\mathrm{e}{-}4\}$; ties within
0.005 go to the lower rate. Validation loss falls monotonically across
the grid for all six models, so every model selects
$2\mathrm{e}{-}4$. For Llama-3.2-1B, the three losses are 1.0002, 0.8490,
and 0.7426. A pre-specified contingency would repeat a pilot at rank 32
if loss failed to improve by 0.02. It never fires.

\paragraph{Gemma-4-E2B corrections.} The reported run is the second of
seven attempts. We record the first attempt because three defects could
each have caused its reversal. First, an unconstrained LoRA target
pattern placed half the adapter on vision and audio components. Second,
the template concatenated prompt and target, taking loss over both
instruction and answer, while generation emitted an unseen empty
reasoning block. Third, shuffled controls ignored language and length.
The repair limits adaptation to 245 language-model projection modules, or
490 LoRA tensors, with no trainable multimodal parameter. It also masks
prompt tokens from the target loss and draws shuffled candidates from the
question's language pool. The reported run passes all three checks.

The pre-repair run also degrades, from 26.4\% to a 23.7\% seed mean, but
it is not a replication. It predates the audit, uses a 1,592-row pooled
denominator instead of Table~\ref{tab:exam}'s per-condition denominator,
and lacks a separate directory of retained per-item outputs. We therefore
cannot recompute it under the reported protocol. It shows only that the
direction appeared before the three defects were repaired.

Within the reported run the seed-42 training crashed near the end on a
learning-rate assertion and was benchmarked from un-converged weights, so
we retrained that seed to its full 216 steps and benchmarked again.

A file-level comparison tests whether that crash explains the result.
Reference, seed-17, and seed-73 files are byte-identical across the two
executions; only seed 42 differs. Yet every strict-consistency cell,
including the seed-42 column, is unchanged. The converged adapter changes
individual predictions but not one rotation-invariant score. These
executions are therefore one experiment with a repaired seed, not two
independent replications.

\paragraph{Adapter scope, before and after the repair.} The two adapters
can be compared directly, and the leak is visible in their weights rather
than only in our description of it. The pre-correction target pattern
matched \texttt{vision}, \texttt{audio\_tower},
\texttt{embed\_vision.embedding\_projection}, and
\texttt{vision\_tower.patch\_embedder} alongside the language
projections. It adapted 494 modules, of which 249 were multimodal. The
corrected pattern matches only language and text branches and adapts 245
modules, none multimodal, which is the 490 LoRA tensors reported above.
Half of the first adapter's capacity was spent outside the language
model.

\paragraph{Training runs on this model.} Gemma-4-E2B took seven training
attempts, more than any other model. Table~\ref{tab:gemmaruns} lists the
ones that recorded metrics. Only the second is reported in this paper;
the rest are training diagnostics, none was benchmarked, and none
contributes a number to any other table.

\begin{table}[ht]
\centering
\small
\setlength{\tabcolsep}{4pt}
\begin{tabular}{llrrrr}
\toprule
Run & LR & step & train & eval & gap \\
\midrule
v20 reported & $2\mathrm{e}{-}4$ & 216 & 0.132 & 0.875 & $+$0.743 \\
v4 gap stop  & $2\mathrm{e}{-}4$ &  54 & 0.129 & 0.990 & $+$0.861 \\
v9 gap stop  & $1\mathrm{e}{-}5$ &  30 & 0.571 & 3.967 & $+$3.396 \\
\bottomrule
\end{tabular}
\caption{Gemma-4-E2B training runs that recorded metrics. Seed means over
three seeds. Only the first is benchmarked or reported elsewhere.
Recomputed by \texttt{gemma\_run\_ablation.py}.}
\label{tab:gemmaruns}
\end{table}

Three further attempts left no metrics: the first, superseded run, whose
adapter config carries the multimodal leak described above; one with no
submitted notebook; and one that crashed on a learning-rate
assertion. A seventh run trained at $1\mathrm{e}{-}5$ under a
relative-gap rule and reached step 216 for two seeds before we cancelled
it with the gap still near 1.4. Only its notebook was archived, so we
report its existence and not its losses.

\paragraph{What the diagnostics show.} At $2\mathrm{e}{-}4$, the gap
reaches 0.86 within 54 steps, one quarter of training. The
$1\mathrm{e}{-}5$ run instead exposes a problem with the stopping
instrument. It stops at step 30 even though the gap shrinks from $+$4.20
to $+$3.85 to $+$3.45. Gemma begins with validation loss 4.81, so an
absolute-gap threshold exceeds its trigger at the first evaluation
regardless of the trajectory. The stopping rule, not the learning rate,
is at fault. A relative rule should stop only when the gap grows from its
own minimum.

The conclusion is narrow. The default rate over-adapts this model
quickly. We do not know how a gentler rate changes its benchmark score
because no lower-rate adapter completed training and scoring under the
frozen protocol. Substituting a rescued recipe for Gemma would compare
one tuned system with five untuned systems, so we do not do so.

\section{Fine-tuning Set Composition}
\label{app:composition}

\begin{table}[ht]
\centering
\small
\setlength{\tabcolsep}{4pt}
\begin{tabular}{lcccr}
\toprule
Act / schedule & S-hop & Adv. & Bar & Tot. \\
 & EN/BN & EN/BN & EN/BN & \\
\midrule
Penal Code       & 90/127 & 111/84 & 39/39 & 490 \\
Evidence Act     & 66/94  & 109/75 & 40/33 & 417 \\
CrPC             & 66/104 & 28/108 & 59/40 & 405 \\
Limitation Act   & 36/42  & 38/43  & 41/40 & 240 \\
Specific Relief  & 17/66  & 6/68   & 40/20 & 217 \\
CPC              & 27/60  & 13/29  & 29/39 & 197 \\
CPC Sch.\ I      & n/a    & n/a    & 39/40 & 79 \\
CrPC Sch.\ II    & n/a    & n/a    & 40/20 & 60 \\
Limitation Sch.\ I & n/a  & n/a    & 20/40 & 60 \\
\midrule
Total & 302/493 & 305/407 & 347/311 & 2,165 \\
\bottomrule
\end{tabular}
\caption{Fine-tuning set by act, category, and language
(English/Bangla), recomputed from the released file. Acts are named in
full in the text. Schedules are listed separately rather than folded into
their parent acts; only the bar-exam category draws on them.}
\label{tab:composition}
\end{table}

\section{Protocol Chronology and Audit Gates}
\label{app:chronology}

\textbf{The chronology is recorded.} Before opening the benchmark, a
written contract froze the model matrix, prompt, training recipe,
constrained scorer, and statistics; manifests record its digests. An
initial single-order execution exposed option sensitivity. Before the
full six-model comparison, we required all four orders and reran every
reported model. This amendment preceded cross-model comparison. Withheld
sessions gated repeated law and none arms.

\paragraph{Rotation chronology and the canary.} The frozen contract
specified the constrained option-letter scorer but not the all-orders
criterion; that amendment is the CircularEval adoption recorded above.
Each E2 session carried 52 item-language pairs stratified by year and
language from the frozen no-context arm. Across one reference, three adapters, and four
rotations, the session proceeded only when all 208 strict item-system
outcomes matched. The gate targets strict outcomes; small rotation-level
argmax changes from batching are recorded but do not enter any reported
aggregate.

\paragraph{Why rotation instead of seed agreement.} Before scoring, we
considered requiring all three training seeds to answer an item
correctly. That rule would be asymmetric: the reference has one
deterministic run and no seed axis, so only the fine-tuned arm would be
penalized for seed variation. Both arms do face option rotation. The
strict metric therefore requires agreement across rotations, not across
seeds.

\paragraph{The irrelevant control matches retrieval length.} The contract
required irrelevant context matched to retrieval length. The first
implementation ran at roughly one quarter of that length. We rebuilt it
to a median 2,000 characters, against 2,002 for retrieval, and use
retrieval minus this condition (R$-$I) only as sensitivity. Retrieval
minus no context (R$-$N) remains the frozen available contrast.

This is the only post-hoc evaluation condition. It replaces the frozen
protocol's shorter shuffled control with sections from the question's
language pool that appear in neither its BM25 nor dense hits. Eleven of
the 398 language rows still contain at least one section overlapping that
item's retrieval, and the reported R$-$I summary retains them.
Contrasting retrieval with this control
separates relevance from the amount of supplied text. The main text
therefore reports both R$-$I and R$-$N.

\section{Difference in Differences by Retrieval Condition}
\label{app:did}

\begin{table}[ht]
\centering
\scriptsize
\setlength{\tabcolsep}{2.5pt}
\begin{tabular}{lrrrrrr}
\toprule
 & \multicolumn{3}{c}{vs.\ none} & \multicolumn{3}{c}{vs.\ irrelevant} \\
\cmidrule(lr){2-4}\cmidrule(lr){5-7}
Model & BM25 & dense & mean & BM25 & dense & mean \\
\midrule
\multicolumn{7}{l}{\emph{Bangla}} \\
Llama-3.2-1B & $+$3.0 & $+$6.0 & $+$4.5 & $+$2.5 & $+$5.5 & $+$4.0 \\
Llama-3.2-3B & $+$5.9 & $+$10.4 & $+$8.1 & $+$0.7 & $+$5.2 & $+$2.9 \\
Qwen3.5-0.8B & $+$5.2 & $+$4.0 & $+$4.6 & $+$4.4 & $+$3.2 & $+$3.8 \\
Qwen3.5-2B & \m2.2 & \m0.2 & \m1.2 & $+$1.0 & $+$3.0 & $+$2.0 \\
Qwen3.5-4B & $+$1.3 & $+$5.7 & $+$3.5 & $+$3.7 & $+$8.0 & $+$5.9 \\
Gemma-4-E2B & $+$0.2 & \m3.9 & \m1.8 & $+$3.9 & \m0.2 & $+$1.8 \\
\midrule
\multicolumn{7}{l}{\emph{English}} \\
Llama-3.2-1B & $+$7.0 & $+$12.4 & $+$9.7 & $+$7.7 & $+$13.1 & $+$10.4 \\
Llama-3.2-3B & $+$3.9 & $+$0.2 & $+$2.0 & $+$1.7 & \m2.0 & \m0.2 \\
Qwen3.5-0.8B & $+$6.9 & $+$8.5 & $+$7.7 & $+$4.4 & $+$6.0 & $+$5.2 \\
Qwen3.5-2B & \m4.2 & $+$0.5 & \m1.8 & \m0.3 & $+$4.4 & $+$2.0 \\
Qwen3.5-4B & \m4.5 & $+$0.8 & \m1.8 & \m1.8 & $+$3.5 & $+$0.8 \\
Gemma-4-E2B & \m1.8 & \m4.5 & \m3.2 & $+$2.2 & \m0.5 & $+$0.8 \\
\bottomrule
\end{tabular}
\caption{Difference in differences by retrieval condition ($\Delta\Delta$
Accuracy). Values are seed means. \emph{mean} averages the BM25 and dense changes
within each seed and is the column reported in Table~\ref{tab:exam}.}
\label{tab:didretriever}
\end{table}

Table~\ref{tab:exam} averages the BM25 and dense changes within each
seed, matching the pair covered by the pre-specified test family.
Table~\ref{tab:didretriever} separates them. They sometimes disagree. For
Gemma-4-E2B in Bangla, the contrast against no context is $+$0.2
with BM25 and \m3.9 with dense retrieval. For Qwen3.5-4B in English, it
is \m4.5 with BM25 and $+$0.8 with dense retrieval. Choosing only one
retriever would therefore change the number of models that appear to show
context-specific gains. We average to avoid that choice.

\section{Additional Results}
\label{app:addl}

\begin{table}[ht]
\centering
\small
\setlength{\tabcolsep}{5pt}
\begin{tabular}{lrrr}
\toprule
Model & ref.\ & FT mean & change \\
\midrule
Llama-3.2-1B & 57.3 & 26.8 & \m30.5 \\
Llama-3.2-3B & 34.7 & 15.7 & \m19.0 \\
Qwen3.5-0.8B & 29.6 &  5.9 & \m23.7 \\
Qwen3.5-2B   & 23.1 & 15.2 &  \m7.9 \\
Qwen3.5-4B   & 31.7 & 14.6 & \m17.1 \\
Gemma-4-E2B  & 13.6 & 18.4 & $+$4.8 \\
\bottomrule
\end{tabular}
\caption{Position-bias spread on Bangla dense rows, as an absolute accuracy
range. The spread
is the range of plain accuracy across the four positions the gold option
can occupy. The fine-tuned column takes the seed mean per position before
the spread, so it is not the mean of per-seed spreads.}
\label{tab:posbias}
\end{table}

Within each model's pre-specified 12-test family, Holm-significant counts
are 12/12 for Llama-3.2-1B, 3/12 for Llama-3.2-3B, 7/12 for
Qwen3.5-0.8B, 0/12 for Qwen3.5-2B, 2/12 for Qwen3.5-4B, and 0/12 for
Gemma-4-E2B. The released analysis JSON contains per-seed differences,
exact McNemar $p$ values, Holm-adjusted $p$ values, and 10,000-draw
item-clustered bootstrap intervals for all 72 comparisons. Those
intervals measure the fine-tuning difference for one retriever and one
seed.

The R$-$N and R$-$I intervals in Table~\ref{tab:exam} come from the same
per-item rows but are computed separately by
\texttt{bootstrap\_did.py}. The script resamples items and checks each
point estimate against the table before reporting an interval width.
Table~\ref{tab:intervals} supplies the intervals for
Table~\ref{tab:control}. \texttt{bootstrap\_gold.py} jointly resamples
the two language readings of 150 control items and 199 examination items
for the law effect and fine-tuning contrasts.
\texttt{bootstrap\_gold\_withheld.py} applies the same procedure to the
withheld-provision run for the provision and same-act decomposition.

\begin{table*}[t]
\centering
\small
\setlength{\tabcolsep}{5pt}
\begin{tabular}{lccccc}
\toprule
Model & law effect & FT$-$Ref, none & FT$-$Ref, law & control & exam R$-$N \\
\midrule
Llama-3.2-1B & \m1.3 & $+$16.2 & $+$14.4 & \m1.8 & $+$7.1 \\
 & [\m7.3, $+$4.7] & [$+$10.4, $+$22.4] & [$+$7.6, $+$21.3] & [\m9.6, $+$5.8] & [$+$4.1, $+$10.2] \\
Qwen3.5-0.8B & $+$18.7 & $+$13.3 & $+$14.4 & $+$1.1 & $+$6.2 \\
 & [$+$11.3, $+$26.0] & [$+$7.8, $+$19.3] & [$+$8.0, $+$21.1] & [\m7.6, $+$9.8] & [$+$3.2, $+$9.0] \\
Llama-3.2-3B & $+$17.3 & $+$3.1 & $+$2.2 & \m0.9 & $+$5.1 \\
 & [$+$10.0, $+$25.3] & [\m2.2, $+$8.4] & [\m4.7, $+$8.9] & [\m8.4, $+$6.7] & [$+$1.2, $+$9.1] \\
Qwen3.5-2B & $+$19.3 & $+$2.7 & $+$1.8 & \m0.9 & \m1.5 \\
 & [$+$12.0, $+$27.3] & [\m2.0, $+$7.3] & [\m3.8, $+$7.3] & [\m8.2, $+$6.4] & [\m5.2, $+$2.1] \\
Gemma-4-E2B & $+$14.7 & \m11.1 & \m6.0 & $+$5.1 & \m2.5 \\
 & [$+$6.7, $+$22.7] & [\m17.6, \m4.9] & [\m11.3, \m0.7] & [\m2.2, $+$12.4] & [\m5.6, $+$0.6] \\
Qwen3.5-4B & $+$17.3 & \m7.6 & \m2.9 & $+$4.7 & $+$0.8 \\
 & [$+$10.7, $+$24.0] & [\m13.6, \m1.8] & [\m6.9, $+$1.1] & [\m2.0, $+$11.1] & [\m2.5, $+$4.1] \\
\bottomrule
\end{tabular}
\caption{Differences and intervals underlying Table~\ref{tab:control}.
\emph{control}: (FT$-$Ref, law) minus (FT$-$Ref, none). Control columns:
$n = 150$; exam R$-$N: $n = 199$.}
\label{tab:intervals}
\end{table*}

\begin{table}[ht]
\centering
\scriptsize
\setlength{\tabcolsep}{1.5pt}
\begin{tabular}{lccc}
\toprule
Model & provision, Ref & provision, FT & same-act, Ref \\
\midrule
Llama-3.2-1B & $+$4.0 [\m0.7, $+$8.7] & $+$6.2 & \m4.7 [\m10.7, $+$1.3] \\
Qwen3.5-0.8B & $+$8.0 [$+$2.0, $+$14.7] & $+$14.0 & $+$10.7 [$+$4.0, $+$18.0] \\
Llama-3.2-3B & $+$10.0 [$+$3.3, $+$16.7] & $+$14.9 & $+$6.7 [0.0, $+$13.3] \\
Qwen3.5-2B & $+$12.7 [$+$6.7, $+$18.7] & $+$14.9 & $+$6.7 [$+$1.3, $+$12.7] \\
Gemma-4-E2B & $+$15.3 [$+$8.7, $+$22.0] & $+$14.2 & \m0.7 [\m8.0, $+$6.7] \\
Qwen3.5-4B & $+$14.7 [$+$9.3, $+$20.7] & $+$13.8 & $+$2.7 [\m3.3, $+$8.7] \\
\bottomrule
\end{tabular}
\caption{Matched withheld-session decomposition ($\Delta$ Accuracy). \emph{provision}
is law minus withheld; \emph{same-act} is withheld minus none. Reference
columns include 95\% item-clustered intervals. FT is the three-seed mean;
no interval is attached to its descriptive provision column.}
\label{tab:withheld}
\end{table}

\textbf{Percentages conceal the sparsest cells.} Several
strict-consistency cells sit near the floor. Table~\ref{tab:sparse}
therefore reports the item count behind every Table~\ref{tab:exam} cell
at or below 5\%, using the same per-item rows and
\texttt{sparse\_cell\_counts.py}.

\begin{table}[ht]
\centering
\small
\setlength{\tabcolsep}{5pt}
\begin{tabular}{lllr}
\toprule
Model & condition & sys & k/199 \\
\midrule
\multicolumn{4}{l}{\emph{Bangla}} \\
Llama-3.2-1B & none & Ref & 1 (0.5\%) \\
Llama-3.2-1B & BM25 & Ref & 1 (0.5\%) \\
Llama-3.2-1B & irrelevant & Ref & 1 (0.5\%) \\
Llama-3.2-1B & dense & Ref & 4 (2.0\%) \\
Qwen3.5-2B & none & Ref & 5 (2.5\%) \\
Qwen3.5-0.8B & none & Ref & 7 (3.5\%) \\
Qwen3.5-0.8B & none & FT & 8 (4.0\%) \\
Llama-3.2-3B & irrelevant & Ref & 8 (4.0\%) \\
\midrule
\multicolumn{4}{l}{\emph{English}} \\
Llama-3.2-1B & irrelevant & Ref & 8 (4.0\%) \\
\bottomrule
\end{tabular}
\caption{Item counts behind every Table~\ref{tab:exam} cell at or below
5\%, out of 199 items. $k$ counts items answered correctly under all four
rotations. FT counts are the mean over the three seeds. The remaining 87
cells sit above 5\%.}
\label{tab:sparse}
\end{table}

\begin{table}[ht]
\centering
\scriptsize
\setlength{\tabcolsep}{2.5pt}
\begin{tabular}{lrrrrrrr}
\toprule
 & \multicolumn{4}{c}{reference} & \multicolumn{3}{c}{seed-42 gain} \\
\cmidrule(lr){2-5}\cmidrule(lr){6-8}
Model & fmt.\ & exact & lenient & logit & exact & lenient & logit \\
\midrule
Qwen3.5-2B   &  5.8 &  4.0 & 33.2 & 51.0 & $+$50.0 & $+$20.9 & $+$3.0 \\
Llama-3.2-1B &  6.0 &  2.8 & 32.2 & 41.0 & $+$39.7 & $+$10.3 & $+$1.8 \\
Qwen3.5-0.8B & 80.7 & 41.7 & 45.0 & 48.0 &  $+$9.8 &  $+$6.5 & $+$3.5 \\
Gemma-4-E2B  & 80.9 & 45.5 & 45.5 & 58.3 & $+$10.1 & $+$10.1 & \m2.5 \\
Llama-3.2-3B & 81.7 & 45.5 & 54.0 & 52.0 &  $+$9.3 &  $+$0.8 & $+$2.8 \\
Qwen3.5-4B   & 98.7 & 62.6 & 63.1 & 61.8 &   \m0.8 &   \m1.3 & $+$0.0 \\
\bottomrule
\end{tabular}
\caption{One comparison read four ways, ordered by reference format
compliance. Rows are the 398 dense-condition rows at rotation 0. fmt.\ is
the percentage of reference generations satisfying the answer grammar,
exact requires one matching line, lenient accepts a letter anywhere, and
logit is the constrained reading. Values are plain rotation-0 accuracies
against a most-frequent-label baseline of 29.1\%, so they are not
comparable to the four-rotation strict-consistency values in
Tables~\ref{tab:control} and \ref{tab:exam}. Lenient parsing removes much
of the inflation at the two least compliant models but not all of it, and
it leaves the Gemma-4-E2B sign disagreement unchanged.}
\label{tab:fourways}
\end{table}

\textbf{The parser effect persists across seeds.} Seed 42 is not
exceptional. Across seeds 17, 42, and 73,
\texttt{recompute\_lenient.py} gives Gemma-4-E2B parser gains of $+$11.3\%,
$+$10.1\%, and $+$10.8\%; lenient and exact-line parsing agree at each seed.
Constrained scoring instead gives \m1.5\%, \m2.5\%, and \m1.8\%. Qwen3.5-2B
shows the same stable inflation: exact-line gains are $+$52.8\%, $+$50.0\%, and
$+$50.8\%, while constrained gains are $+$5.5\%, $+$3.0\%, and $+$3.8\%.

\textbf{Quoted text does not explain the law effect.} Choosing the option
that most quotes the supplied law scores 56/150 (37.3\%) on the control.
In a reviewer-requested post-hoc check, we remove ties and keep the 92
items where a distractor has strictly greater 5-gram coverage than the
correct option. Reference law effects remain $+$12.0\%, $+$19.6\%, $+$16.3\%,
\m2.2\%, $+$22.8\%, and $+$13.0\% for Qwen3.5-0.8B, Qwen3.5-2B,
Qwen3.5-4B, Llama-3.2-1B, Llama-3.2-3B, and Gemma-4-E2B. The script
\texttt{handbuilt\_gold\_results.py} reconstructs this subset from
recorded contexts, options, and per-item outputs. This analysis was not
pre-specified.

\section{Translation Audit}
\label{app:transaudit}

\begin{table}[ht]
\centering
\small
\setlength{\tabcolsep}{5pt}
\begin{tabular}{lrrr}
\toprule
Planted error & n & judge A & judge B \\
\midrule
key option swapped & 8 & 8 & 2 \\
non-key option replaced & 8 & 8 & 1 \\
legally inert edit (false positive) & 6 & 1 & 0 \\
\bottomrule
\end{tabular}
\caption{Judge sensitivity on mechanically planted corruptions with known
ground truth. Entries count items rated major or critical; for inert
edits a flag is a false positive. Judge A's 1 of 6 there has a 95\%
Wilson interval reaching 56\%.}
\label{tab:judges}
\end{table}

At temperature 0, each judge sees the Bangla and English items side by
side. It returns a severity in \{none, minor, major, critical\}, whether
the answer is preserved, whether the option set changed, whether legal
terms survived, and a free-text note. Major and critical judgments count
as flags. Judge A is \texttt{gemini-3.6-flash}; judge B is
\texttt{gemini-2.5-flash-lite}. Claude Sonnet 4.5 produced the
translations, so neither judge evaluates output from its own model
family.

Among the 177 items that were never seeded, judge A flags 25 (14.1\%,
95\% CI 9.8\% to 20.0\%) and preserves the recorded answer for 168 (94.9\%).
Judge B flags 3 of 176 (1.7\%, 95\% CI 0.6\% to 4.9\%) and preserves 175
answers. After restoring seeded items to clean text, A flags 29 of all
199 items; B flags 4 of the 198 items for which it returns a verdict.

\begin{table}[ht]
\centering
\small
\setlength{\tabcolsep}{4pt}
\begin{tabular}{lrrrrr}
\toprule
 & \multicolumn{3}{c}{all items} & \multicolumn{2}{c}{by year} \\
\cmidrule(lr){2-4}\cmidrule(lr){5-6}
Model & 199 & $-$8 & $-$25 & 2022 & 2023 \\
\midrule
Llama-3.2-1B & $+$9.7 & $+$10.5 & $+$10.2 & $+$7.6 & $+$11.8 \\
Llama-3.2-3B & $+$2.0 & $+$1.3 & $+$4.1 & $+$0.3 & $+$3.7 \\
Qwen3.5-0.8B & $+$7.7 & $+$7.9 & $+$9.4 & $+$5.6 & $+$9.8 \\
Qwen3.5-2B & \m1.8 & \m1.7 & \m1.6 & \m2.2 & \m1.5 \\
Qwen3.5-4B & \m1.8 & \m1.6 & \m2.2 & \m4.9 & $+$1.2 \\
Gemma-4-E2B & \m3.2 & \m3.3 & \m3.4 & \m3.0 & \m3.3 \\
\midrule
\multicolumn{4}{l}{items flagged by judge A} & 29/99 & 0/100 \\
\bottomrule
\end{tabular}
\caption{English difference in differences on benchmark subsets
($\Delta\Delta$ Accuracy). The contrast is that of Table~\ref{tab:exam}, recomputed from the
raw per-item rows. $-$8 drops the items judge A called answer-changing
and $-$25 every never-seeded item it flagged, leaving 191 and 174. The
last two columns split by examination year. All 29 flags lie in 2022.}
\label{tab:subsets}
\end{table}

\section{Translation Robustness}
\label{app:transrobust}

The translation explanation fails two checks in
Table~\ref{tab:subsets}. Removing the 8 answer-changing items moves every
English difference in differences by at most 0.8\%, and removing all 25
flagged items by at most 2.1\%; neither flips a sign. Translation noise
should also
inflate the year containing all 29 flags, but 2022 is the smaller year
for five of six models.

\section{Reproducibility}
\label{app:repro}

\textbf{Availability.} The corpus, QA dataset, benchmark, per-item
outputs, and analysis code are public under permissive terms; adapter
weights are available on request. They are described in enough detail
throughout this appendix that a reader can judge what exists without
following a link. The anonymized review copy is available at
\url{https://anonymous.4open.science/r/bangladesh-legal-qa-11E3}; it
exposes the contract, corpus, QA dataset, benchmark, translation audit,
free-form generations, training-run registry, manifests, and analysis
scripts. The run tree contains six reported checkpoint executions, two
superseded executions retained for audit, and a separate Gemini rotation
diagnostic that uses a different scorer. Repository and dataset URLs that
identify the authors are withheld until acceptance. Licensing is stated
in the Ethics Statement.

From the repository root,
\texttt{python 07-analysis/scripts/reproduce.py} runs the 21 bundled
analyses in about one minute without a GPU or network. Six additional
repository checks read \LaTeX{} sources or adapter weights and therefore
sit outside the release command. Hard gates fail on mismatches; soft
checks print values for manual comparison. The command
\texttt{python 07-analysis/scripts/reproduce.py -{}-list} prints the
claim map connecting every script to its claim and regenerated folder.

The release includes the structured bilingual corpus, the 2,165-record QA
set and its pre-review pool, the group-disjoint split and digest, the
frozen retrieval bundle, the experiment contract, the adapter manifests,
per-item outputs for the examination and original supplied-law runs, raw
free-form generations, Gemma training notebooks, run manifests, and
analysis scripts. Each benchmark directory records the model revision,
prompt and contract hashes, scorer, package versions, hardware, peak
memory, and each result file's SHA-256. This supports numerical
reproducibility. It does not imply complete provenance for every archived
run; two known gaps appear below. Superseded checkpoint runs and the
external Gemini diagnostic remain visible because they audit how the
reported protocol was reached; neither enters the six-checkpoint tables.

\textbf{The manifests make the freeze checkable.} Readers can verify that
one contract and one prompt governed the eight checkpoint executions, six
reported and two superseded. Their manifests record
\hashv{419b73b8b1b33e0a8298b9bd15801cad2c87bd9b07ea62af7ac5ee9eb3ef8f02},
and the released contract hashes to that value. The same manifests carry
one prompt digest,
\hashv{ec2c0196c3db4879612433e528f34d1c2050ebb13656a4535436741c12629d18},
across every run. Editing the contract after observing a result would
break these matches. Hashes alone do not establish calendar order;
repository history provides that record.

An independent script also re-audits the five non-Gemma models. It
reparses all 40 result files, confirms every recorded hash, and checks
199 item identifiers, two languages, four conditions, four rotations, no
duplicate composite keys, no cancelled items, and identical keys between
each fine-tuned run and its reference. Only then does it recompute every
reported statistic across all 127,360 constrained rows.

Two provenance gaps remain. The archived Qwen3.5-0.8B and Qwen3.5-2B
benchmark directories lack the \texttt{RUN\_COMPLETE} marker file,
although both carry terminal manifests with end times and all eight
result hashes verified. And for the corrected Gemma-4-E2B adapters the
recomputed local directory-tree hashes do not reproduce the tree hashes
recorded in the training manifests, even though the file inventory
matches and every \texttt{safetensors} file is complete; the individual
weight-file SHA-256 values are therefore the durable provenance for that
model, and the benchmark outputs produced with those adapters are
manifest-hash verified.

\end{document}